\documentclass[letterpaper, 10 pt, conference]{ieeeconf}  

\IEEEoverridecommandlockouts                              

\usepackage[dvipsnames,table,xcdraw]{xcolor} 
\usepackage{cite} 
\usepackage{dsfont} 

\makeatletter
\let\NAT@parse\undefined
\makeatother

\usepackage[colorlinks=true,
            citecolor=CornflowerBlue,
            linkcolor=black,
            urlcolor=orange]{hyperref}

\usepackage[utf8]{inputenc}
\usepackage[T1]{fontenc}
\usepackage[english]{babel}
\usepackage{fontawesome5}
\usepackage{tabularx}
\usepackage{lipsum}
\usepackage{hhline}
\usepackage{bm}
\usepackage{pdfpages}
\usepackage{acro}
\usepackage{bookmark}
\usepackage{etoolbox}
\usepackage{wrapfig}
\usepackage{stmaryrd}
\usepackage{rotating}
\usepackage{mwe}
\usepackage{multicol}
\usepackage{multirow}
\usepackage{lettrine}
\usepackage{tikz}
\usetikzlibrary{calc}
\usepackage{environ}
\usepackage{adjustbox}
\usepackage{siunitx}
\usepackage{pifont}
\usepackage{mathtools}
\usepackage{amsmath,amssymb}
\usepackage{mdframed}
\usepackage{pgfplots}
\pgfplotsset{compat = newest}
\usepackage{booktabs}
\usepackage{tcolorbox}
\usepackage{fontawesome5}

\definecolor{lightgray}{HTML}{EFEFEF}
\definecolor{midgray}{HTML}{C0C0C0}
\definecolor{darkgray}{HTML}{9B9B9B}
\definecolor{coolred}{HTML}{E77475}
\definecolor{coolblue}{HTML}{277C9D}
\definecolor{coolgreen}{HTML}{598938}
\definecolor{coolyellow}{HTML}{FACB77}
\definecolor{coolorange}{HTML}{FF8C00}
\definecolor{coolcyan}{HTML}{3EC3B2}
\definecolor{coollightblue}{HTML}{62BCDD}
\definecolor{coolpurple}{HTML}{976AA3}
\definecolor{ColComments}{HTML}{F7C5A6}

\definecolor{softpurple}{HTML}{EDE7F6}
\definecolor{softblue}{HTML}{E6F0FA}
\definecolor{softgreen}{HTML}{E8F5E9}
\definecolor{softgrayblue}{HTML}{EEF2F7}

\newcommand{\myrule}{\specialrule{1pt}{0pt}{0pt}}
\newcolumntype{M}[1]{>{\centering\arraybackslash}m{#1}}
\newcolumntype{Y}{>{\centering\arraybackslash}p}
\newcolumntype{Z}{>{\centering\arraybackslash}X}

\newcommand{\eg}[0]{\textit{e.g.},\ }

\newcommand{\ie}[0]{\textit{i.e.}\ }

\newcommand{\ours}[0]{Ours}
\newcommand{\contm}[0]{\textit{\ours{} w/ context=$\hat{\mathbf{m}}_t$}}
\newcommand{\contz}[0]{\textit{\ours{} w/ context=$\mathbf{0}$}}

\newcommand{\hc}[0]{\# HC \textbf{↓}}
\newcommand{\tc}[0]{\# EC \textbf{↓}}

\newcommand{\myparagraph}[1]{\noindent \textbf{#1}}

\title{\LARGE \bf
Advantage-Driven Explicit Memory for Social Navigation
}

\author{Yeonsoo Park$^{1}$\quad Mattia Racca$^{2}$\quad Guillaume Bono$^{2}$\quad Steeven Janny$^{2}$ \\ Gianluca Monaci$^{2}$\quad Tomi Silander$^{2}$\quad Christian Wolf$^{2}$
\thanks{$^{1}$ Yeonsoo Park is with the Interdisciplinary Program in Artificial Intelligence, Seoul National University, Korea; the work was done while at Naver Labs Europe ({\tt\small yeonsoopark33@gmail.com}). $^{2}$ authors are with Naver Labs Europe, France
        ({\tt\small \{firstname.lastname\}@naverlabs.com})}%
}

\makeatletter
\let\NAT@parse\undefined
\makeatother
\begin{document}

\maketitle
\thispagestyle{plain}    
\pagestyle{plain}         

\begin{abstract}
Robot policies are predominantly learned with classical parametric variants of imitation learning or RL, where training stores the agent's behavior exclusively in the policy's network parameters,  putting a heavy burden on the representation learning algorithm. We propose a new navigation agent equipped with non-parametric memory which explicitly indexes prior steps leading to critical events. The advantages are twofold: first, it allows the policy to outsource some of its behavior into an explicit memory; second, it encourages a form of continual learning by allowing an agent to collect data from its testing episodes during deployment and therefore to better generalize to  OOD situations. In the context of social navigation, we show that this improves the agent's capability to retain sparse, high-cost failures, such as human collisions. If the policy is trained in simulation, this also naturally addresses the sim-to-real gap, partially, by basing some of the decision making on real data.
We integrate the explicit memory into a recurrent PPO architecture and use hidden states for memory retrieval to capture continuous spatiotemporal dynamics. The goal of exploiting rare, high-impact events is achieved by leveraging the RL agent's advantage signals. We train our agent in simulation with a combination of photorealistic rendering and non-visual crowd simulation and show that the agent is robust with respect to OOD social behavior.
\end{abstract}

\section{Introduction}

\noindent
Safe and efficient navigation in highly interactive, human-populated environments remains a fundamental challenge in robotics, primarily due to the unpredictable nature of human motion and the difficulty of resolving rare edge cases, such as navigating around blind corners or negotiating the use of a narrow doorway.
While recent advancements have heavily relied on parametric learning with Imitation Learning (IL) or Reinforcement Learning (RL) to handle complex situations, these standard approaches heavily rely on the power of 
representation learning to compress all necessary agent behavior into the weights of the policy's neural network. By averaging past experiences into implicit network weights, crucial high-error signals from challenging navigation moments risk being diluted within the parameter space.
Without the ability to explicitly anchor past critical experiences in some form of memory, agents tend to struggle to flexibly adapt to specific mistakes, often repeating unsafe behaviors in out-of-distribution scenarios.

Inspired by retrieval-augmented generation (RAG), explicit memories have been proposed recently for agents ~\cite{BlundellX16ModelFreeEpisodicControl,LinIJCAI18EpisodicMemoryDeepQNetworks,DouNIPS19FastDRLOnlineAdjustments,he_ma-lmm_2024} and in particular for navigation \cite{GoyalICML22RetrievalAugmentedRL,xie2024embodiedraggeneralnonparametricembodied,monaci2025rana}, with the goal of partially delegating reasoning to external memory and lightening the burden of the algorithm training the policy weights. Non-parametric memories provide a powerful complement to parametric learning by allowing agents to directly store and retrieve past experiences, injecting explicit supplementary context to guide the reasoning process. 
Despite this potential, existing memory-based approaches in navigation typically store snapshot observations or spatial embeddings, \eg \cite{GoyalICML22RetrievalAugmentedRL,xie2024embodiedraggeneralnonparametricembodied,monaci2025rana}.
Navigating human-populated spaces, however, requires understanding the continuous cause-and-effect of interactions, rather than merely retrieving isolated moments in time.

\begin{figure}
    \centering
    \includegraphics[width=\linewidth]{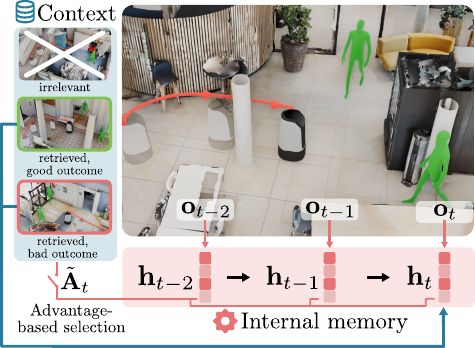}
    \caption{We address human-aware social navigation through explicit memory. During navigation, our end-to-end agent collects and stores past experiences in its context buffer, and can search through it at each step in order to find similar situations and their corresponding outcomes. Context push and pulls are based on advantage and value estimation, effectively filling the context with the most unexpected situations.}
    \label{fig:teaser}
\end{figure}

We propose a method that allows retrieval-augmented agents to exploit \textit{event memories}, capturing the full spatiotemporal history of challenging interactions. It centers on action-outcome integration, explicitly linking the agent's actions to their delayed consequences. This is achieved by filling the agent's explicit memory with stored agent states, each obtained from a given time instant $t$ and linked to the future outcome of the agent from this state onward, calculated as its return, \ie the exact cumulated future rewards. 
This structures the agent's history as a situational cause-and-effect ``Scorecard'' and enables the robot to query utility metrics derived from past experiences. 

Beyond the gains of partially delegating agent behavior to a retrieval process, this mechanism also naturally allows to address distribution shifts between training and test environments, particularly useful in social navigation and in settings where training is performed in simulation. As part of the agent behavior stems from the exploitation of its explicit memory, we show that filling this memory with test interactions can lead to significant gains in OOD situations.

Our main contributions are as follows:

\begin{itemize}
    \item \textbf{Retrieval from agent states:} by using recurrent hidden agent states as memory keys, our method retrieves experiences based on continuous interaction dynamics rather than instantaneous spatial snapshots. This enables the agent to extract and utilize past interactions that are aligned with its current situation.
    \item \textbf{Rare event preservation:}     
    we utilize large-magnitude advantage signals to trigger the storage of rare, high-impact experiences. This is in contrast to classical RL training, where sparse training signals from critical failures tend to be diluted. This can also be seen as a form of inductive bias, which forces the preservation of high-magnitude (positive or negative) advantage signals and counters their catastrophic forgetting.
    \item \textbf{Action-outcome event logs:} 
    we structure memory as a ``Scorecard'', which explicitly maps candidate hidden states and actions to their resulting utilities. By integrating this direct action-outcome evaluation into an RL/PPO architecture, the agent can assess delayed consequences of actions directly,
    improving sample efficiency during training.
    \item \textbf{Online adaptation to OOD situations at test time:} we propose a strategy for collecting and scoring memory items directly within the deployment environment, which enables on-the-fly behavioral adaptation without requiring additional gradient updates. By substituting the train memory buffer with testing experiences, this property effectively mitigates distribution shifts.
\end{itemize}

\section{Related Works}

\myparagraph{Robot Navigation} has been classically addressed using explicit modeling and considering humans as any other obstacle~\cite{burgard1998interactive,macenski2020marathon}.
More recently, end-to-end trained navigation models directly map input to actions and are typically trained with RL~\cite{mirowski17learning,bono2024learning} or IL~\cite{DBLP:conf/nips/DingFAP19}.
Our approach, inspired by~\cite{janny2025}, adopts a recurrent policy trained end-to-end with RL in simulation, using a motion model based on dynamics identified from real trajectories and quantized velocity commands to reduce the need for advanced low-level control and improve trajectory smoothness when deploying on real robots.

Enabling robots to navigate among humans is a natural extension of end-to-end trained agents \cite{mavrogiannis23,francis25}, with humans introducing expectations, intentions, and social norms, requiring robots to account not only for safety and efficiency but also for socially appropriate behavior.
However, this setting gives rise to rare but high-cost events, such as collisions with humans at blind corners or bottlenecks, which pose challenges for  learning approaches.
The proposed advantage-driven explicit memory addresses this issue by storing rare experiences and making them available for future decision-making through retrieval.

\myparagraph{Retrieval-augmented RL.}
Non-parametric memories store past experiences in an external dataset rather than internalizing them into model parameters. Episodic control methods for RL~\cite{BlundellX16ModelFreeEpisodicControl,LinIJCAI18EpisodicMemoryDeepQNetworks,DouNIPS19FastDRLOnlineAdjustments,he_ma-lmm_2024} 
act on successful experiences by
re-employing Q-value estimates, 
enhancing sample efficiency during training.
In contrast,
we 
exploit retrieved information from past navigation experiences to preserve rare but important events and enhance navigation.
\textit{Goyal et al.}~\cite{GoyalICML22RetrievalAugmentedRL}
augment an RL agent with retrieval parameterized as a neural network that has access to a dataset of past trajectories.
\cite{humphreys_largescalererl} proposes an RL agent
enhanced with a non-parametric retrieval mechanism based on pre-trained features.
\textit{RANa} \cite{monaci2025rana} exploits a dataset of unposed visual observations collected by one or more robots and uses semantic and geometric foundations models for retrieval and for extracting directional information. Here instead we propose a retrieval augmentation for an RL agent using the current state as a query to retrieve relevant past trajectories from the memory buffer. 

\myparagraph{Memory-Based Approaches in Navigation.} 
Existing memory-augmented navigation methods are based on occupancy maps~\cite{Chaplot2020Learning}, semantic maps~\cite{chaplot2020object}, latent metric maps~\cite{DBLP:conf/pkdd/BeechingD0020,DBLP:conf/iclr/ParisottoS18}, topological maps~\cite{BeechingECCV2020,shah2022viking}, multi-modal hierarchical graphs~\cite{xie2024embodiedraggeneralnonparametricembodied}, explicit episodic memory~\cite{Fang_2019_CVPR,reed_generalist_2022}, implicit recurrent episodic memory \cite{kinaema2025}, object-centered implicit representations trained at test-time ~\cite{Marza2022NERF} and image collections~\cite{monaci2025rana}. However, such representations are insufficient for social navigation, which requires understanding complex and prolonged human interactions.
Our work addresses this limitation by storing and retrieving recurrent hidden states to capture temporal dynamics and long-term interactions.
\begin{figure*}[!ht]
    \centering
    \includegraphics[width=\textwidth]{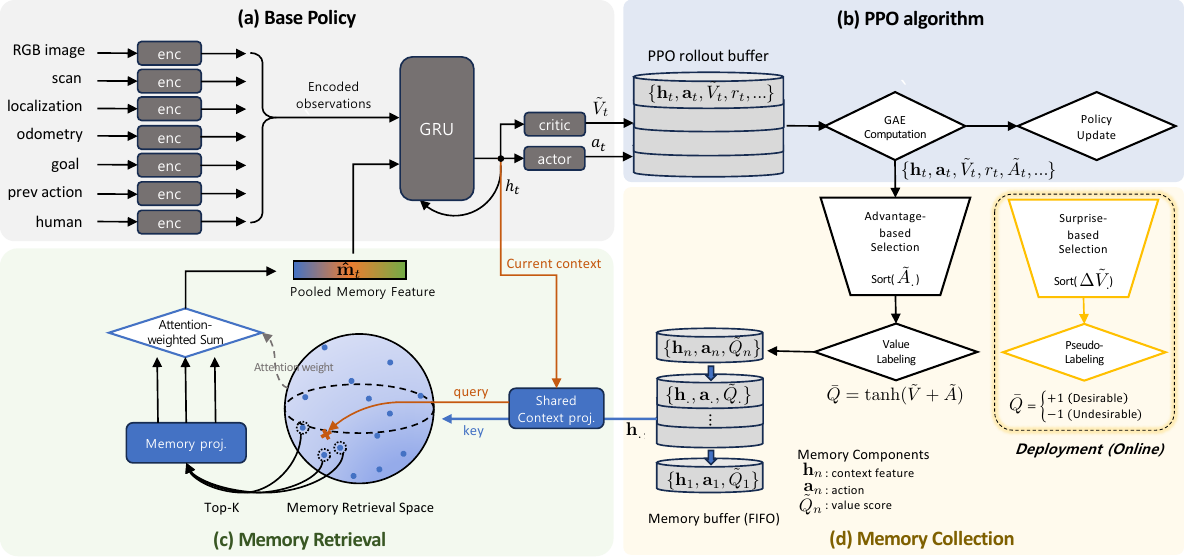}
    \caption{The architecture of the augmented human-aware navigation agent. 
    \textbf{(a) Base Policy:} A GRU-based recurrent network processes multi-modal observations ($\mathbf{o}_t$) along with the retrieved pooled memory feature ($\mathbf{\hat{m}}_t$) to generate the navigation action ($a_t$). 
    \textbf{(b) PPO Algorithm:} classically, a rollout buffer is managed, and advantage is calculated.
    \textbf{(c) Memory Retrieval:} The current context ($\mathbf{h}_{t-1}$) and stored memory contexts are mapped into a shared Memory Retrieval Space. Memory slot features ($\mathbf{m}_n$) are dynamically aggregated via an attention-weighted sum based on the cosine similarity between the query ($\mathbf{h}_{t-1}$) and keys ($\mathbf{h}_n$). 
    \textbf{(d) Memory Collection:} Rollout experiences undergo GAE computation to evaluate their utility. The method employs Advantage-based Selection for offline training and Surprise-based Selection for online deployment. Selected experiences are assigned a bounded value score ($\bar{Q}_t$) and stored as episodic tuples in the FIFO memory buffer.}
    \label{fig:overall_architecture}
\end{figure*}

\section{A memory-augmented RL agent}

\noindent
Our agent is based on the end-to-end trained agent from \textit{Janny et al.}~\cite{janny2025}, which at each time step $t$ receives a set of observations $\mathbf{o}_t$ including: a RGB image captured from an onboard camera, a LiDAR-like vector of ranges, and two different localization signals (odometry- and mapping-based).
At the beginning of every navigation episode, the agent is given a static goal vector (in 2D polar coordinates), and both localization signals are reset, \ie they are expressed in an episodic coordinates frame relative to the agent initial position and orientation.
The agent outputs a 2D velocity command $a_t=(a_v, a_w)$ as a pair of linear and angular velocities
selected from a discrete set of $28$ possible actions: $4$ possible linear velocities combined with $7$ possible angular velocities.
The different inputs go through dedicated encoders: a half-width ResNet for the image, a 1D-CNN for the scan, a discrete embedding for the previous action, and dedicated small Multi-Layer Perceptrons (MLPs) for all other inputs.

Similarly to a large body of work \cite{bono2024learning,janny2025,DBLP:conf/iclr/JaderbergMCSLSK17,khanna2024goatbench,bongratz2024chooseRLAlg,CrocoNav2024}, the agent maintains a latent memory vector $\mathbf{h}_t$ defined as the hidden state of a Gated Recurrent Unit (GRU),
which gets updated before each decision based on received observations
\begin{equation}
    \mathbf{h}_t = \mathrm{GRU}(\mathbf{h}_{t-1}, \mathrm{enc}(\mathbf{o}_t))\text{ .}
    \label{eq:gru}
\end{equation}
Finally, based on the latent memory vector, two independent linear layers, the actor ($\pi$) and the critic ($\phi$) heads,
produce the distribution over actions from which $a_t$ will be sampled and an estimate of the current state value $\tilde{V}_t$, respectively:
\begin{align}
    a_t \sim & \pi(\cdot | \mathbf{h}_t)\text{ ,}
    \label{eq:policy} \\
    \tilde{V}_t = & \phi(\mathbf{h}_t)\text{ .}
    \label{eq:critic}
\end{align}
In addition to these standard observations, our social navigation task implies interacting with humans.
To this end, the agent also receives at each step a short trace of the position of surrounding humans in its field of view,
and we augment the agent of \cite{janny2025} with a ``human encoder'':
a transformer-based module inspired by \textit{Wayformer} \cite{wayformer2022} first extracting features of the trace of each individual person,
then combining them together to form global crowd features,
which gets concatenated to the rest of the observations features and given as input to the GRU.
We also add a $29^\text{th}$ action, a duplicated $(0,0)$ command to enable the agent to differentiate between stopping at the goal and waiting for humans to move.
The whole agent is fine-tuned end-to-end in simulation with RL starting from the weights of~\cite{janny2025}, with Proximal Policy Optimization (PPO)~\cite{schulman2017proximal}, maximizing the return induced by the standard dense navigation reward function
\begin{equation}
    r = R_\text{succ}.\mathds{1}_\text{succ}
    - R_\text{dist}\Delta\text{dist}
    - C_\text{colli}.\mathds{1}_\text{colli}
    - C_\text{slack}
\end{equation}
where $R_\text{succ}=2.5$ is the gain on successful episode,
$R_\text{dist}=1.0$ is the gain on reducing length of shortest path towards goal,
$C_\text{colli}=0.1$ is the cost on collision,
and $C_\text{slack}=0.01$ is the slack cost.

While recurrent networks can implicitly encode interaction history, updating their parameters relies purely on gradient learning, which often fails to reliably preserve rare but critical interaction patterns~\cite{bengio1994learning}. To address this, we introduce an external episodic memory module directly integrated into a PPO-GRU pipeline. This allows the agent to explicitly store and retrieve critical spatiotemporal events, enabling continuous behavioral adaptation. We first describe the overall architecture in Section \ref{architecture}, followed by the memory retrieval and fusion mechanism in Section \ref{memory_fusion}, and conclude with our causal episodic indexing strategies for both training and deployment phases in Section \ref{entri}.

\subsection{Memory Augmentation}
\label{architecture}
\noindent The proposed method augments the recurrent agent with an external, non-parametric episodic memory module. Figure \ref{fig:overall_architecture} illustrates the overall architecture, which operates through three primary components: the recurrent base policy, a spatiotemporal memory retrieval mechanism, and a causal memory collection module.

In addition to the embeddings of the observations, we propose to concatenate an additional vector of memory features $\mathbf{\hat{m}}_t$ to the input of the GRU.
The hidden state $\mathbf{h}_{t-1}$ from the previous step, which encodes the temporally-integrated interaction history, is explicitly utilized as the current spatiotemporal context to query a memory buffer $\mathcal{M} = \{ (\mathbf{h}_n, \mathbf{m}_n)\ :  n\in\{0\dots N\} \}$ accumulated over past navigation episodes. We define
\begin{equation}
    \mathbf{\hat{m}}_t = \sum_{n=0}^N \alpha_{t,n}\ \mathbf{m}_n\text{ ,}
\end{equation}
where $\alpha_{t, n}$ are weighting ratios between the last internal state $\mathbf{h}_{t-1}$ and the states stored in memory buffer $\{\mathbf{h}_n: n\in\{0\dots N\}\}$,
and $\{\mathbf{m}_n: n\in\{0\dots N\}\}$ are entries in the memory buffer built from internal state, action and expected future value for selected steps of past navigation episodes.
In the current implementation, the memory buffer has a fixed capacity $N=512$, and behaves as a ``First-In-First-Out'' (FIFO) queue.
Future work could explore different discard mechanisms for memory entries.

\subsection{Spatiotemporal Memory Retrieval and Fusion}
\label{memory_fusion}

\noindent
This section describes the mechanism for retrieving and integrating historical interaction data into the current decision-making process. The method utilizes a non-parametric memory bank $\mathcal{M} = \{ (\mathbf{h}_n, \mathbf{m}_n)\ : n\in\{0\dots N\} \}$, where each entry $\mathbf{m}_n$ is a vector built from internal state $\mathbf{h}_n$, the executed action $a_n$, and saturated future Q-value $\bar{Q}_n \in [-1, 1]$ (cf. Section~\ref{entri}) from a selected step $t_n$ in a past navigation episode.

\subsubsection{Memory Structure and Querying}
Rather than using instantaneous observations, we query memory through the recurrent hidden state $\mathbf{h}_{t-1} \in \mathbb{R}^d$ of the GRU.
This state serves as both the query for the current retrieval and the key for future memory storage.
To evaluate the relevance of past experiences, we project the current context $\mathbf{h}_{t-1}$ and the stored memory contexts $\mathbf{h}_n$ into a shared retrieval space, using a two-layer MLP $f_c$, then compute their cosine similarity as
\begin{equation}
    \sigma_{t,n} = \frac{f_c(\mathbf{h}_{t-1}) \cdot f_c(\mathbf{h}_n)}{||f_c(\mathbf{h}_{t-1})||\ ||f_c(\mathbf{h}_n)||}\text{ .}
\end{equation}
$f_c$ is introduced to adapt the pre-trained GRU to the new retrieval objective (with orthogonality regularization), while preserving the existing latent state representation.

To further enhance the discriminative power of the retrieval space and prevent representation collapse, we augment the PPO loss with an orthogonal regularization term
\begin{equation}
    \sum_{\mathbf{h}\neq \mathbf{h'}} \underbrace{\bar{f}_c(\mathbf{h}) \cdot \bar{f}_c(\mathbf{h'})^\top}_\text{Outer prod.} - \mathbf{I}
\end{equation}
for all internal states $\mathbf{h}$ and $\mathbf{h'}$ collected in the rollout buffer during trajectories sampling with current policy,
where $\bar{f}_c$ denotes the normalized projection, and $\mathbf{I}$ the identity matrix.

\subsubsection{Retrieval and Scorecard Aggregation}
The retrieval mechanism selects the top-$K$ relevant experiences (with $K=5$) according to the computed similarities $\sigma_{t,n}$, which are then normalized using softmax with temperature scaling $\tau=1.0$
\begin{align}
    \alpha_{t,n} =& \frac{\exp(\sigma_{t,n} / \tau)}{\sum_{n'} \exp(\sigma_{t,n'} / \tau)}\text{ ,}
\end{align}
where $\sigma_{t,n_k}$ is masked out (i.e. set to $-\infty$) if $n_k$ is not part of the top-$K$ selection.

\subsection{Construction of Causal Episodic Entries}
\label{entri}

\noindent
This section describes how memory entries $\mathbf{m}_n$ are built and which steps $t_n$ are selected to populate the memory buffer, using a \emph{$k$-step look-ahead} mechanism.

\subsubsection{Training phase: Offline Selection and Labeling}

At the end of every navigation episode, we compute an estimate of the advantage $\tilde{A}_t$ of the executed action $a_t$ given internal state $\mathbf{h}_t$ for the current policy $\pi$ at every decision step $t$ using Generalized Advantage Estimation (GAE)~\cite{schulman2015high}
\begin{equation}
    \tilde{A}_t = \underbrace{r_t + \gamma \tilde{V}_{t+1} - \tilde{V}_t}_\text{critic TD error} + \underbrace{\lambda\gamma\tilde{A}_{t+1}}_\text{GAE exp. avg.}\text{ ,}
\end{equation}
where $r_t$ is the instantaneous reward returned by the environment, $\gamma=0.99$ the discount factor applied to future values, $\lambda=0.95$ the averaging factor used by GAE.

We use the absolute value $|\tilde{A}_t|$ of the advantage as an indicator of how ``surprising'' the outcome of a given action was, in both directions, \ie decisions which yielded either better or worse outcomes than expected compared to just following current policy.
During training, after collecting a batch of episodes, we use this absolute advantage indicator to partition all collected steps and keep the top-$p$\% (with $p=5$\%) to be added to the memory buffer.
For all steps $t_n$ belonging to this top-$p$ partition, we compute memory entry $m_n$ as
\begin{equation}
    m_n = f_m(f_c(\mathbf{h}_{t_n - k})\ ..\ \mathbf{E}[a_{t_n - k}]\ ..\ f_v(\bar{Q}_{t_n}))\text{ ,}
\end{equation}
where $f_m$ and $f_v$ are learned linear projections, $..$ denotes concatenation, $\mathbf{E}$ is a discrete set of learned action embedding vectors,
and  $\bar{Q}_t=\tanh(\tilde{V}_t + \tilde{A}_t)$ is a saturated Q-value estimate based on GAE.

Note the $k=6$ offset on the decision step for both the internal state and action.
When a high-impact event is detected at time $t_n$ according to $|\tilde{A}_{t_n}|$,
storing the instantaneous state $\mathbf{h}_{t_n}$ provides limited utility for preventive learning, as the state already reflects the expected outcome.
To capture the causal precursors of these events, we instead index the spatiotemporal context and the executed action from $k$ steps prior.
This temporal shift allows the agent to retrieve the corresponding ``Scorecard'' early enough during future rollouts to execute preemptive maneuvers before a similar situation materializes. 
This $m_n$ entry is associated with $\mathbf{h}_n = \mathbf{h}_{t_n - k}$ in the memory buffer.

The saturated Q-value $\bar{Q}_{t_n}$ represents how good or bad the interaction was in the episode final outcome.
Squashing this signal into the $[-1, 1]$ range prevents numerical instability and allows the attention mechanism to interpret the score intuitively: $+1$ signals a behavior to be emulated, while $-1$ indicates an action to be strictly avoided.

\subsubsection{Deployment phase: Online Selection and Labeling}
\label{online-selection}

In deployment scenarios, the offline-constructed memory buffer may face inherent limitations in Out-of-Distribution (OOD) environments where social dynamics or obstacle configurations deviate significantly from the training distribution. To bridge this gap, we introduce a training-free online collection strategy that enables rapid adaptation.
A critical challenge in OOD scenarios is the potential for hallucination in the state-value network (Critic); since the Critic has not encountered such states, its value estimations often become unreliable, leading to erroneous memory labeling.

As we do not have access to all the privileged information (environment state) to compute the same reward function as in training, 
we replace our ``surprise'' indicator based on GAE by either one of the following heuristics to feed the memory buffer:
\begin{itemize}
    \item Use absolute temporal difference of values predicted by the learned critic $|\tilde{V}_{t+1} - \tilde{V}_t|$ with no instantaneous reward.
    \item Use a sparse reward function $r'$ such that $r'_t = 0$ almost everywhere along the trajectory except for final reached states where
        agent receives $r_T=+1$ if the episode was successful (goal reached in time without collision)
        and $r_T=-1$ otherwise (collided, aborted far from goal, or exhausted steps budget)
\end{itemize}

This mechanism effectively injects ``cheat sheets'' into the agent's working memory for current environment and out-of-distribution (OOD) scenarios.
Furthermore, our method supports human-curated indexing, allowing a human supervisor to manually select and label specific interaction points to override or reinforce behaviors. Because the value scores are saturated through $\tanh$ squashing, an expert can intuitively assign labels ($+1$ for good, $-1$ for bad) to override or reinforce behaviors. This flexibility allows the agent to adapt to novel real-world dynamics without requiring backpropagation. 

\subsubsection{Memory Maintenance and Sustainability}
Since the encoder is jointly trained with the policy, representation drift may affect memory retrieval over long training horizons. However, our system mitigates this issue through a combination of FIFO memory replacement and similarity-based retrieval filtering, which naturally suppress outdated memory entries.

\section{Experiments}

\subsection{Simulation Environment and Base Agent Setup}

\noindent
We conduct our experiments in the Habitat-Sim environment~\cite{puig2023habitat3} utilizing the Habitat Matterport 3D (HM3D)~\cite{ramakrishnan2021hm3d} dataset for high-fidelity indoor navigation.
The task is formulated as PointGoal navigation \cite{DBLP:journals/corr/abs-1807-06757} in crowded dynamic environments, with the crowd being simulated through the Recast Navigation library \cite{recastnavigation}.
As in the SocNavBench~\cite{biswas22}, the crowd does not directly react to the agent, thereby representing a challenging scenario in which people do not adapt to the robot’s navigation behavior and may collide with it if it obstructs their path.
This design choice prevents the trained agent from excessively relying on the crowd’s implicit collision avoidance. 
Our agent utilizes a GRU as the policy body, processing multi-modal inputs: LiDAR scans, front RGB camera, odometry, localization, goal coordinates, previous actions, and the 2D poses of surrounding humans.

\subsubsection{Training Environment and Curriculum}
During training, human agents are programmed to move between randomly sampled waypoints, with velocity and acceleration of crowd members sampled from a normal distribution informed by the data collection effort of~\cite{drazen14}.

We employ a curriculum learning strategy based on crowd density: the maximum density $\rho_{max}$ is progressively increased from $0.05$ to $0.2$ people per square meter.
Note that during training, the density for each episode is uniformly sampled from $[0, \rho_{max}]$, so that, at regime (when $\rho_{max} = 0.2$), the average crowd density is $0.1$.

\subsubsection{Evaluation Methodology}
To evaluate the robustness and adaptability of the proposed framework, we define two testing scenarios based on crowd complexity and density:

\begin{itemize}
    \item \textbf{In-Distribution (ID):} This scenario maintains the configuration of the final training stage. Crowd movements follow the random waypoint model, and the crowd density is uniformly sampled at each episode from the range $[0, \rho_{max}]$ with $\rho_{max} = 0.2$. This represents the environment the agent was optimized for at the end of the curriculum learning, with an expected density of $0.1$.

    \item \textbf{Out-of-Distribution (OOD):} This scenario introduces substantial distributional shifts through the following changes to the crowd dynamics:
    \begin{itemize}
        \item \textbf{High Density:} The crowd density is held constant at $\rho = 0.2$ for every episode. This removes the lower-density scenarios encountered during training, resulting in a consistently congested environment.
        \item \textbf{Novel Movement Patterns:} Crowd members now exhibit behaviors beyond the random waypoint wandering of the training phase, namely \textit{grouping} (synchronized cluster movement) and \textit{idling} (standing still in navigable paths).
        \item \textbf{Wilder Crowds:} The velocity distribution is modified to produce highly divergent (very slow and very fast) samples. Furthermore, we increased the variance of the acceleration to incorporate more erratic and aggressive movement profiles.
    \end{itemize}
\end{itemize}
PointGoal navigation performance is quantified using standard metrics: Success Rate (SR), Success weighted by Path Length (SPL), and Success weighted by Completion Time (SCT).
Safety is measured by the total number of environment collisions (\# EC) and collisions with humans (\# HC).

\subsection{The impact of humans on baseline performance}
\begin{table}[t]
    \centering
    \caption{Impact of evaluation with humans on the agent without explicit memory: success rate drops as the agent is disturbed by human presence.}\label{tab:eval_baseline}
    \begin{tabular}{c|>{\cellcolor{softblue}}c>{\cellcolor{softgreen}}c}   
        \multicolumn{3}{c}{Backbone\cite{janny2025} w/ human enc, trained w/ crowd} \\ 
        \myrule
        \rowcolor{midgray}
        \cellcolor{lightgray}  & \cellcolor{softblue!200} Eval w/o crowd   &\cellcolor{softgreen!200} Eval w/ crowd \\ \myrule
         SR (\%) \textbf{↑}    & \textbf{89.48}   &  76.16 \\
        SPL (\%) \textbf{↑}    & \textbf{76.75}   &  65.65 \\
        SCT (\%) \textbf{↑}    & \textbf{32.35}   &  28.50 \\
        \hc{}                  & n.a.\            &  2.67\\
        \tc{}                  & 6.15             &  \textbf{5.35}\\ \myrule
    \end{tabular}
\end{table}

\noindent
Table \ref{tab:eval_baseline} evaluates the model from \cite{janny2025} augmented with human encoders only, but without explicit memory, trained with crowds, in two different settings: with and without crowds.
The baseline agent indeed suffers from the presence of the crowd compared to the empty environment, despite being trained with crowds.

\subsection{Performance on In-Distribution Scenarios}
\begin{table}[t]
    \centering
    \caption{Impact of explicit memory and retrieval, in-distribution (same crowd behavior in train and eval): augmented agent gains in success rate while keeping human collisions low.}\label{id_performance}
    \begin{tabular}{c|>{\cellcolor{softblue}}c>{\cellcolor{softpurple}}c>{\cellcolor{softgreen}}c}
        \rowcolor{white}\multicolumn{4}{c}{Backbone \cite{janny2025} w/ human encoders, trained w/ crowd} \\         
        \myrule
        \cellcolor{lightgray} &\multicolumn{2}{c}{\cellcolor{lightgray}No explicit memory} &\cellcolor{softgreen!200}Explicit memory \\    
        \cellcolor{lightgray} &\cellcolor{softblue!200}baseline & \cellcolor{softpurple!200}context=$\mathbf{0}$ & \cellcolor{softgreen!200}context=$\mathbf{\hat{m}}_t$ \\
        \myrule
         SR (\%) \textbf{↑}          & 76.16    & 79.60           & \textbf{81.28}  {\color{coolgreen}(+1.68)}  \\
        SPL (\%) \textbf{↑}         & 65.65    & 69.48           & \textbf{70.33}    {\color{coolgreen}(+0.85)}      \\
        SCT (\%) \textbf{↑}          & 28.50    & 29.56           & \textbf{30.45}    {\color{coolgreen}(+0.89)}      \\
        \hc{} & \textbf{2.67}   & 3.12          & 3.33     {\color{coolred}(+0.21)}    \\
        \tc{} & 5.35   & \textbf{5.18}          & 5.21      {\color{coolred}(+0.03)}   \\
        \myrule
    \end{tabular}
\end{table}

\noindent
Table \ref{id_performance} summarizes the performance under the standard training distribution (ID).
To verify the contribution of our training strategy and the episodic memory, we compare the baseline PPO-GRU against our framework in two configurations: without the memory buffer (\contz), where the memory feature is set to zero, and with the full trained buffer (\contm).
In particular, we compare against \contz{} to verify that the performance gains do not simply emerge from the additional network capacity and fine-tuning.

The results show that our framework outperforms the baseline even when the external memory is inactive.
Specifically, \contz{} achieves an SR of 79.60\%, a 3.44\% absolute improvement over the baseline (76.16\%).
This increase shows that our advantage-driven memory selection strategy during training 
increases sample efficiency by forcing the base policy to ``lock on'' to critical interaction patterns.
Unlike standard PPO, which treats all roll-out samples with equal priority, our method prioritizes high-impact events (\eg near-collisions), allowing the base weights to internalize social navigation logic during the training phase itself.

When the trained memory buffer is activated (\contm), the SR further increases to 81.28\%. 
Since the trained buffer is populated with experiences collected during the training phase, these stored memories closely align with the dynamics of the ID test environment.
This consistent gain across all metrics demonstrates that the agent successfully performs real-time reasoning by retrieving these highly relevant historical ``Scorecards.''
The performance scaling between the \contz{} and \contm{} settings proves that the memory module does not simply provide redundant information; rather, it provides a useful reference that allows the agent to navigate high-density situations that exceed the reactive capacity of a standard recurrent policy.

\subsection{Training-free Online Adaptation in OOD Environments}
\label{exp_ood}
\begin{table}[t]
    \centering
    \caption{Impact of explicit memory and retrieval, out-of-distribution (different social behavior between train and eval): performance drops further due to the distribution shift, but the augmented agent can provide significant gains by leveraging test time samples}
    \label{tab:ood_results}
    \begin{tabular}{c|>{\cellcolor{softblue}}c>{\cellcolor{softpurple}}c>{\cellcolor{softgreen}}c}
        \rowcolor{white}\multicolumn{4}{c}{Backbone \cite{janny2025} w/ human encoders, trained w/ crowd} \\         
        \myrule
        \cellcolor{lightgray} &\multicolumn{2}{c}{\cellcolor{lightgray}No explicit memory} &\cellcolor{softgreen!200}Explicit memory \\    
        \cellcolor{lightgray} &\cellcolor{softblue!200}baseline & \cellcolor{softpurple!200}context=$\mathbf{0}$ & \cellcolor{softgreen!200}context=$\mathbf{\hat{m}}_t$ \\
        \myrule
         SR (\%) \textbf{↑}          & 54.00    & 56.40                  & \textbf{60.00} {\color{coolgreen}(+3.60)}  \\
        SPL (\%) \textbf{↑}         & 46.92    & 48.39                  & \textbf{52.16} {\color{coolgreen}(+3.77)} \\
        SCT (\%) \textbf{↑}         & 20.41    & 20.72                  & \textbf{23.12} {\color{coolgreen}(+2.40)}  \\
        \hc & 3.32     & 3.64                   & \textcolor{white}{-}\textbf{3.24} {\color{coolgreen}(-0.40)}   \\
        \tc & 4.26     & \textbf{3.78}          & \textcolor{white}{-}3.84 {\color{coolred}(+0.06)}    \\ \myrule
    \end{tabular}
\end{table}

\noindent
Table \ref{tab:ood_results} summarizes the performance comparisons under the aforementioned OOD scenario, with fixed high density and novel social behaviors.

\subsubsection{The Limits of Offline Memory}
As expected, the severe distributional shift causes the baseline PPO-GRU's performance to degrade significantly, yielding a Success Rate (SR) of only 54.00\%.
Relying solely on the offline \textit{trained buffer} provides a marginal improvement (SR 56.40\%).
This limited gain confirms our hypothesis: when the stored ``Scorecards'' represent training dynamics that no longer align with the underlying OOD environment, the static offline memory struggles to provide adequate guidance.

\subsubsection{Training-Free Online Adaptation}
To bridge this gap, we evaluate the \textit{online buffer} configuration, where the agent dynamically constructs memory entries on-the-fly using the physical heuristic-based selection described in Section \ref{entri}.
As shown in Table \ref{tab:ood_results}, employing the online buffer yields a substantial performance recovery, boosting the SR to 60.00\% (+3.60\% over the trained buffer) and the SPL to 52.16\%.
Notably, the human collision metric drops to 3.24, indicating that the agent successfully utilizes newly injected negative labels ($\bar{Q}_t = -1$) from recent physical outcomes to rapidly suppress catastrophic failures in unseen social contexts.
This demonstrates that our framework provides incremental, immediate behavioral benefits without requiring any backpropagation.

\subsubsection{Scalability}
A common critique of memory-augmented architectures is the inference latency caused by querying large buffers.
In our experiments however, as memory capacity increases from 512 to 4096 entries, the average query time remains strictly bounded between 0.10 ms and 0.125 ms, with no significant latency degradation.
This suggests that our framework can support long-term knowledge accumulation without compromising the real-time reactive control required for safe physical deployment.

\begin{table}
    \caption{Ablation Study on Query Features and Orthogonal Loss for Memory Retrieval}
    \label{abl1}
    \begin{tabular}{c|>{\cellcolor{softblue}}c>{\cellcolor{softpurple}}c>{\cellcolor{softgreen}}c>{\cellcolor{softgrayblue}}c}
        \myrule
        \cellcolor{lightgray}& \cellcolor{softblue!200}\textbf{Hid. State} &\cellcolor{softpurple!200} \textbf{Hid. State} &\cellcolor{softgreen!200} \textbf{All Obs}         &\cellcolor{softgrayblue!200} \textbf{Human}      \\
        \cellcolor{lightgray}& \cellcolor{softblue!200}$+L_\text{orth}$ &\cellcolor{softpurple!200} &\cellcolor{softgreen!200}  &\cellcolor{softgrayblue!200}\textbf{feats}      \\
        \myrule
         SR (\%) \textbf{↑}  & \textbf{81.28}      & 77.00        & 73.84           & 73.72           \\
        SPL (\%) \textbf{↑}  & \textbf{70.33}      & 66.49        & 64.68           & 64.30           \\
        SCT (\%) \textbf{↑}  & \textbf{30.45}      & 28.23        & 26.84           & 25.31           \\
        \hc                  & 3.33              & 2.73       & 2.51          & \textbf{2.24} \\
        \tc                  & 5.21              & 5.15       & \textbf{4.38} & 4.61          \\ \myrule
    \end{tabular}
\end{table}
\begin{table}
    \caption{Ablation Study on Online Buffer Value Scoring Strategies}
    \label{abl2}
    \begin{tabular}{l|>{\cellcolor{softblue}}c>{\cellcolor{softpurple}}c>{\cellcolor{softgreen}}c>{\cellcolor{softgrayblue}}c}
        \cellcolor{lightgray} &  \cellcolor{softblue!200} Trained & \multicolumn{3}{c}{\cellcolor{lightgray}Online Buffer} \\
        \cellcolor{lightgray}  & \cellcolor{softblue!200} Buffer & \cellcolor{softpurple!200}Ours          &\cellcolor{softgreen!200} Pred.\ value        & \cellcolor{softgrayblue!200}$\operatorname{sgn}(\text{R})$     \\ \myrule
        SR (\%) \textbf{↑}           & 56.40          & \textbf{60.00} {\color{coolgreen}(+3.6)} & 59.60 {\color{coolgreen}(+3.2)}      & 56.40 {\color{gray}(+0.0)}          \\
        SPL (\%) \textbf{↑}          & 48.39          & \textbf{52.16} {\color{coolgreen}(+3.8)} & 51.45 {\color{coolgreen}(+3.1)}      & 49.12 {\color{coolgreen}(+0.7)}          \\
        SCT (\%) \textbf{↑}          & 20.72          & \textbf{23.12} {\color{coolgreen}(+2.4)} & 22.42 {\color{coolgreen}(+1.7)}      & 21.78 {\color{coolgreen}(+1.0)}          \\
        \hc & 3.64           & 3.24 {\color{coolgreen}(-0.40)}           & 4.00 {\color{coolred}(+0.36)}      & \textbf{3.23} {\color{coolgreen}(-0.41)}  \\
        \tc & 3.78           & 3.84 {\color{coolred}(+0.06)}           & 4.10 {\color{coolred}(+0.32)}      & \textbf{3.70} {\color{coolgreen}(-0.08)}  \\ \hline
    \end{tabular}
\end{table}
\subsection{Ablations}
\subsubsection{Memory Query Features}

Table \ref{abl1} evaluates different context features used as memory retrieval queries, revealing that the query design strongly shapes the agent's navigation strategy.
Following standard practices, we utilize the GRU hidden state as the baseline context.
The configuration augmenting the hidden state with orthogonal loss achieves the highest navigation performance, yielding an SR of 81.28\%.
This confirms that the orthogonal loss effectively disentangles the metric space, preventing representation collapse and thereby increasing memory coverage.
In contrast, querying with all concatenated raw observations (All Obs) minimizes total static collisions but significantly degrades the Success Rate to 73.84\%, as the inclusion of excessive sensory noise forces the agent into an overly conservative policy.
Relying exclusively on the output of the human encoder (Human Feat) transforms the agent into an ``avoidance specialist'', with the lowest number of human collisions (2.24) but a significant drop in navigation performance (SR 73.72\%). 
The choice of memory query therefore determines the agent’s focus and resulting strategy, providing an additional design lever to tailor the agent’s behavior to the requirements of the environment.

\subsubsection{Online Labeling Strategies}
Table \ref{abl2} evaluates the impact of value scoring strategies during online memory collection, following the experimental setup of Section \ref{exp_ood}.
Relying on the critic's predicted value (Pred.\ value) proves unreliable in OOD environments, causing human collisions to spike to 4.00.
Using just the sign of the reward~(\texttt{sgn}(R)) decreases collisions but stagnates success rate improvements.
We therefore resort to a rule-based labeling based on physical outcomes (see Section~\ref{online-selection}): we deterministically assign $-1$ for critical failures (\eg human collisions) and $+1$ for successful navigation.
These labels act as a crude proxy for human preference, and they appear to balance social safety and high navigation efficiency.

\section{Conclusion and Discussion}

\noindent
In this work we introduced a novel non-parametric episodic memory used to augment an RL-trained navigation agent with the objective to address the inherent limitations of parametric reinforcement learning in out-of-distribution social navigation. By integrating a $k$-step look-ahead mechanism, our approach explicitly captures the causal precursors of high-impact events.
Beyond strict navigation performance, the method offers three distinct advantages regarding training efficiency and real-world deployment. First, it reliably preserves rare but critical interaction patterns. In standard RL, sparse signals from critical failures (\eg collisions) produce minute gradients that are easily diluted. Episodic memory, however, is frequency-independent; it persistently retains high-cost events, ensuring the agent continuously references these isolated failures during decision-making. Second, it significantly improves sample efficiency by accelerating credit assignment. By explicitly storing the direct causal linkage between early contextual actions and their delayed consequences, our agent bypasses the long-horizon bottlenecks of backpropagation through time. Third, it inherently mitigates the sim-to-real gap. While the classical agent's policy is entirely frozen by sim-trained parameters, our non-parametric buffer can dynamically accumulate real-world interaction memories, which enables training-free, on-the-fly adaptation during physical deployment. 

Future work will explore scalable buffer management strategies for lifelong learning and investigate more robust online selection criteria. Furthermore, while accumulating real-world entries mitigates behavioral gaps, the latent query space itself (the sim-trained hidden state) may still experience a sim-to-real gap. Utilizing vision foundation models to extract domain-agnostic retrieval queries presents a promising avenue to fully close this gap. Furthermore, we are currently working on testing this agent in real environments with humans detected by a state-of-the-art mesh recovery model \cite{anny2025}. More challengingly, the model could be fed not only the human positions and orientations but the full body pose described by the \textit{Anny} mesh \cite{anny2025}, as well as face features, which would require additional learning from real social data sequences.



\bibliographystyle{IEEEtran}
\bibliography{bib,main_cvpr2024,main_cvpr2025,main_eccv2025}

\end{document}